\documentclass[10pt,twocolumn,letterpaper]{article}

\usepackage{iccv}
\usepackage{times}
\usepackage{epsfig}
\usepackage{graphicx}
\usepackage{amsmath}
\usepackage{amssymb}
\usepackage{epsfig}
\usepackage{graphicx}
\usepackage{amsmath}
\usepackage{amssymb}
\usepackage{algorithm}
\usepackage{algorithmic}
\usepackage[switch]{lineno}
\usepackage{subcaption}
\usepackage{makecell, multirow, tabularx}
\usepackage{booktabs}
\usepackage{cite}
\usepackage{color}
\usepackage{amsmath}
\definecolor{hollywoodcerise}{rgb}{0.96, 0.0, 0.63}
\definecolor{lasallegreen}{rgb}{0.03, 0.47, 0.19}
\definecolor{hanpurple}{rgb}{0.32, 0.09, 0.98}
\definecolor{green(pigment)}{rgb}{0.0, 0.65, 0.31}
\usepackage{amssymb}% http://ctan.org/pkg/amssymb
\usepackage{pifont}% http://ctan.org/pkg/pifont
\newcommand{\cmark}{\ding{51}}%
\newcommand{\xmark}{\ding{55}}%
\usepackage{amsmath}
\newcommand*{\affmark}[1][*]{\textsuperscript{#1}}
\definecolor{hollywoodcerise}{rgb}{0.96, 0.0, 0.63}
\definecolor{lasallegreen}{rgb}{0.03, 0.47, 0.19}
\definecolor{hanpurple}{rgb}{0.32, 0.09, 0.98}
\definecolor{green(pigment)}{rgb}{0.0, 0.65, 0.31}
\iccvfinalcopy % *** Uncomment this line for the final submission
\def\iccvPaperID{****} % *** Enter the ICCV Paper ID here

\ificcvfinal\fi

\begin{document}

%%%%%%%%% TITLE
\title{A Good Student is Cooperative and Reliable: CNN-Transformer Collaborative Learning for Semantic Segmentation}

\author{Jinjing Zhu$^{1}$
\quad
Yunhao Luo$^{3}$
\quad
Xu Zheng$^{1}$
\quad
Hao Wang $^{4}$
\quad
Lin Wang$^{1,2}$ \thanks{Corresponding author}
\and
\affmark[1] AI Thrust, HKUST(GZ)\quad
\affmark[2] Dept. of CSE, HKUST \quad
\affmark[3] Brown University\quad
\affmark[4] Alibaba Cloud, Alibaba Group\\
\quad
{\tt\footnotesize zhujinjing.hkust@gmail.com, devinluo27@gmail.com, zhengxu128@gmail.com,
cashenry@126.com, linwang@ust.hk}}

\ificcvfinal\thispagestyle{empty}\fi
\twocolumn[{
\renewcommand\twocolumn[1][]{#1}%
\captionsetup{font=small}
\maketitle
\ificcvfinal\thispagestyle{empty}\fi
\captionsetup{font=small}
\begin{center}
\vspace{-19pt}
    \centering
    \includegraphics[ width=0.99\textwidth, height=4cm]{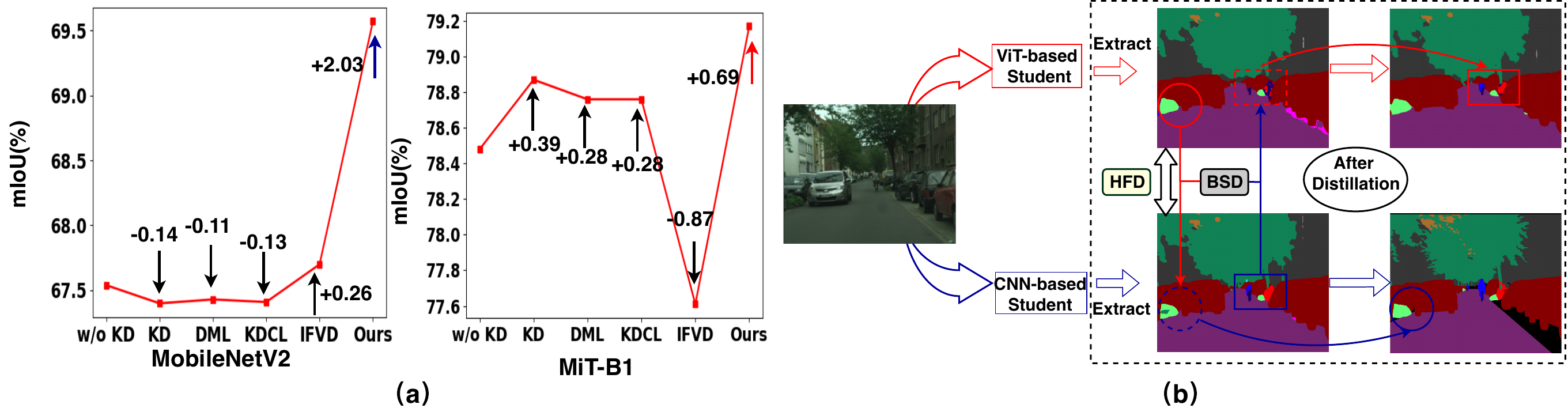}
\vspace{-7pt}
    \captionof{figure}{(a) Our CNN-ViT collaborative learning framework can learn the compact ViT-models and CNN-based models simultaneously while achieving the SoTA segmentation performance than prior methods. (b) We propose the first online KD framework to collaboratively learn compact CNN-based and ViT-based models by selecting and exchanging reliable knowledge between them. }
\label{fig:Introduction}
\vspace{-5pt}
\end{center}}] 
%%%%%%%%% ABSTRACT
\begin{abstract}
\vspace{-6pt}
   In this paper, we strive to answer the question `\textit{how to collaboratively learn convolutional neural network (CNN)-based and vision transformer (ViT)-based models by selecting and exchanging the reliable knowledge between them for semantic segmentation?}' Accordingly, we propose an online knowledge distillation (KD) framework that can simultaneously learn compact yet effective CNN-based and ViT-based models with two key technical breakthroughs to take full advantage of CNNs and ViT while compensating their limitations. Firstly, we propose heterogeneous feature distillation (\textbf{HFD}) to improve students' consistency in low-layer feature space by mimicking heterogeneous features between CNNs and ViT. Secondly, to facilitate the two students to learn reliable knowledge from each other, we propose bidirectional selective distillation (\textbf{BSD}) that can dynamically transfer selective knowledge. This is achieved by 1) region-wise BSD determining the directions of knowledge transferred between the corresponding regions in the feature space and 2) pixel-wise BSD discerning which of the prediction knowledge to be transferred in the logit space. Extensive experiments on three benchmark datasets demonstrate that our proposed framework outperforms the state-of-the-art online distillation methods by a large margin, and shows its efficacy in learning collaboratively between ViT-based and CNN-based models.
\end{abstract}
\section{Introduction}
Semantic segmentation \cite{LongSD15, XieWYAAL21, ChenPKMY18} is a crucial and challenging vision task, which aims to predict a category label for each pixel in the input image. Although the state-of-the-art (SoTA) segmentation methods have achieved remarkable performance, they often require prohibitive computational costs. This limits their applications to resource-limited scenarios, \eg, autonomous driving \cite{FengHRHGTWD21}. Consequently, growing attention has been paid to model compression aiming at obtaining more compact networks. It can be roughly divided into quantization \cite{WuLWHC16, bs-2103-13630, EsserMACAABMMBN16}, pruning \cite{CaiAYYX22, LouizosWK18, abs-1803-05729}, and knowledge distillation (KD)~\cite{WangY22,SonNCH21, ParkCJKH21}. The standard KD paradigm aims to learn a compact yet effective student model under the guidance of a high-capacity teacher model. For instance, CD \cite{ShuLGYS21} proposes a channel-wise KD approach by normalizing the activation map of each channel. IFVD \cite{WangZJBX20} characterizes the intra-class feature variation (IFV) and makes the student model mimic the IFV of the teacher model. 

Recently, vision transformer (ViT) achieves comparable or even better performance than that of CNNs thanks to the computing paradigm, \eg, multi-head self-attention (MHSA). For instance, PVT~\cite{WangX0FSLL0021, WangXLFSLLLS22} and Swin Transformer \cite{abs-2111-09883, DongBCZYYCG22} extract the pyramid features from the high-resolution images and achieve SoTA performance on various benchmarks. To minimize the model complexity, SegFormer~\cite{XieWYAAL21} proposes a hierarchically structured transformer encoder to learn a simple yet efficient ViT-based model. 

In this paper, we strive to collaboratively learn \textit{compact} yet \textit{effective} CNN-based and ViT-based models for semantic segmentation. Intuitively, we explore an online KD paradigm for this goal. Existing online KD methods for classification employ a `Dual-Student' framework (without the pre-trained model) by enabling the students to learn from each other in a one-stage learning manner \cite{AnilPPODH18, SongC18, ZhangXHL18, GuoWWYLHL20}. For example, Deep Mutual Learning (DML) \cite{ZhangXHL18} proposes to make the CNN-based students teach each other in the training process. KDCL~\cite{GuoWWYLHL20} enables the students with different capacities to learn collaboratively to generate reliable soft supervision and boost their classification performance. However, naively applying these CNN-based KD methods is less effective and even leads to performance drops (see Fig.~\ref{fig:Introduction} (a)).
The reasons are that: 1) 
% Online KD ignores the performance gap between CNN and ViT [ViT and CNN difference]. For instance, CNNs are hardcoded to attend only locally and ViT does not learn to attend locally in earlier layers;
The discrepancies in the feature and prediction space between CNNs and ViT caused by the distinct computing paradigms make it challenging to perform online KD.
2)
% These methods only transfer knowledge in the logits space, which lacks effective and reliable guidance in the feature space because ViT [].
These methods only transfer knowledge in the logit space while more reliable and informative knowledge does exist in the \textit{feature space}.
3) There are considerable model size gap and learning capacity gap between CNNs and ViT. 
Intuitively, we ask a question: \textit{`how to collaboratively learn CNN-based and ViT-based models by selecting and exchanging the reliable knowledge between them for semantic segmentation?'
% How to explore a collaborative learning strategy by selecting or exchanging the reliable knowledge between the ViT-based and CNN-based students for semantic segmentation?'
}

In light of this, we propose, to the best of our knowledge, the \textbf{first} online KD strategy 
to further push the limit of CNNs and ViT for semantic segmentation (See Fig. \ref{fig:Introduction} (b)). Our method enjoys two key technical breakthroughs.
Firstly, we propose heterogeneous feature distillation (\textbf{HFD}) to make the students learn the heterogeneous features from each other for complementary knowledge in the low-layer feature space. Concretely, the ViT-based student takes the low-level features from the CNN-based student as guidance and vice versa.
% perform self-attention in ViT-based student with the lower-layer features of the CNN-based student as input. [goal of this module, the why do we propose, []
% Then, we align the output with low-layer features of ViT-based students and can enable CNN-based student to learn global feature representations. 
% Similarly,  we perform convolution in CNN-based student with the lower-layer features of the ViT-based student as input. 
%Then we align the output with low-layer features of CNN-based student. 
Then, consistency between the low-layer features of CNN-based and ViT-based students is imposed to encourage them to compensate for their limitations.
Secondly, to transfer reliable knowledge between CNNs and ViT, we propose a bidirectional selective distillation (\textbf{BSD}) module that selectively distills the reliable region-wise and pixel-wise knowledge.
Specifically, the region-wise distillation dynamically transfers reliable knowledge of regions in the feature space by determining the directions of transferring knowledge. Similarly, pixel-wise distillation discerns which of the prediction knowledge to be transferred in the logit space. Note that these bidirectional distillation approaches are both guided by the cross entropy between predictions and ground-truth (GT) labels. 

In summary, our main contributions are four-fold: (\textbf{I}) We introduce the \textit{first} online collaborative learning strategy to collaboratively learn compact ViT-based and CNN-based models for semantic segmentation. (\textbf{II}) We propose HFD to facilitate CNNs and ViT learning global and local feature representations correspondingly. (\textbf{III}) We propose BSD to distill knowledge between ViT and CNNs in the feature and logit spaces. (\textbf{IV}) Our proposed method consistently achieves new state-of-the-art performance on three benchmark datasets for semantic segmentation.

\begin{figure*}[t]
    \centering
    \includegraphics[width=0.99\textwidth, height=7.5cm]{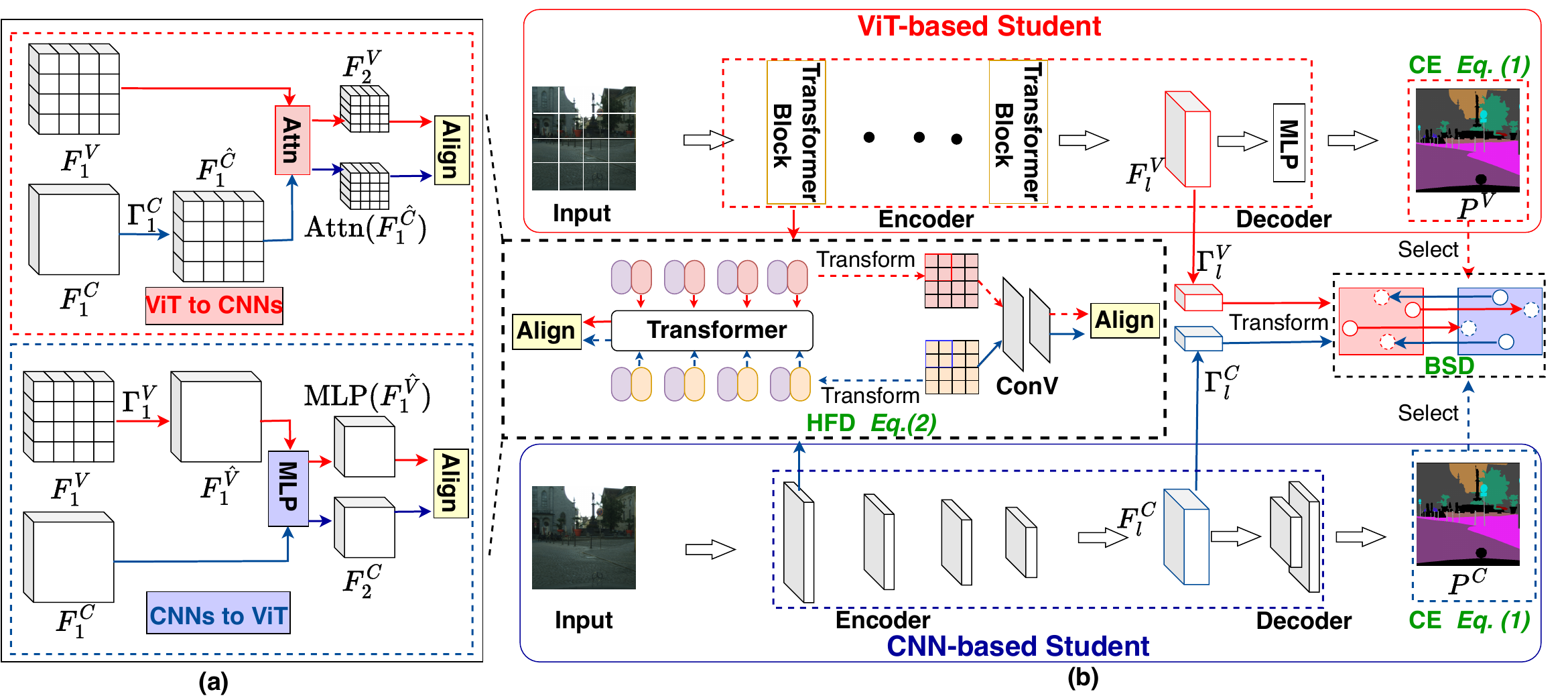}
    % \vspace{-8pt}
    \caption{Illustration of the proposed framework, containing three parts: ViT-based student, CNN-based student, and KD modules. (a) Heterogeneous feature distillation (HFD) enables CNN-based and ViT-based students to learn from each other in the low-layer feature space. (b) Our proposed framework. The online KD strategy is optimized via three loss functions: (i) a cross entropy loss, (ii) heterogeneous feature distillation loss, and (iii) a bidirectional selective distillation (BSD) loss.}
    \label{fig:famework}
   \vspace{-10pt}
\end{figure*}

\section{Related work}

\noindent \textbf{KD for Segmentation.} 
% Semantic segmentation aims to make pixel-wise predictions that can be utilized in a large number of real-world applications, such as autonomous driving \cite{WangCY20} and metaverse \cite{abs-2110-05352}. 
% To minimize the computational costs of segmentation networks,
% Various works  have been proposed for obtaining compact segmentation models by using KD techniques. 
The mainstream methods~\cite{WangZJBX20, LiuCLQLW19, JiWTH0L22, ShuLGYS21, HeSTGSY19, AnilPPODH18, YaoS20, KimHCK20, BeyerZRMA022} for segmentation mostly focus on learning a compact CNN-based student model by distilling the knowledge from a cumbersome CNN-based teacher model with the same network architecture. 
CD \cite{ShuLGYS21} proposes a channel-wise KD approach by normalizing the activation map of each channel. IFVD \cite{WangZJBX20} characterizes the intra-class feature variation (IFV) and makes the student mimic the IFV of the teacher. SSTKD \cite{ji2022structural} exploits the structural and statistical knowledge to enrich low-level information of the student model. \textit{Differently, we propose a first online KD approach, which collaboratively learns compact yet effective CNN-based and ViT-based models for segmentation.}

\noindent \textbf{Vision Transformer} has demonstrated its effectiveness on several vision tasks but is less applicable in case of limited computational resources. Recently, several attempts have been made to obtain compact ViT models via network pruning~\cite{abs-2204-07154} or KD~\cite{JiaoYSJCL0L20}. Moreover, some works \cite{abs-2108-05895} combine the advantages of CNNs and ViT, and design hybrid models for classification.
\textit{By contrast, we explore simultaneously learning compact yet effective CNNs and ViT models by bidirectionally learning the feature and prediction information from both models for semantic segmentation.
}
% To demonstrate the effectiveness of our method, we utilize the CNNs (MobileNet \cite{HowardZCKWWAA17} and ResNet \cite{he2016deep}) and ViT (SegFormer \cite{XieWYAAL21}) to assign labels of pixels in images.  

% model compression \cite{abs-2204-07154} or knowledge distillation \cite{JiaoYSJCL0L20} techniques to obtain compact ViT models. Moreover, some works \cite{abs-2108-05895} design hybrid light-weight models by leveraging the advantages of CNNs and ViT. In this work, we propose a collaborative learning strategy CTCL to make CNNs and ViT learn from each other. To demonstrate the effectiveness of our method, we utilize the CNNs (MobileNet \cite{HowardZCKWWAA17} and ResNet \cite{he2016deep}) and ViT (SegFormer \cite{XieWYAAL21}) to assign labels of pixels in images.  

\noindent \textbf{Online KD.} Some works \cite{ZhangXHL18, AnilPPODH18, ChenMWF020, KimHCK20} focus on the online KD without a pre-trained teacher model. DML~\cite{ZhangXHL18} proposes a mutual learning strategy, where an ensemble of students' logits is deployed, for classification task. Co-distillation \cite{AnilPPODH18} further extends this idea and explores the potential in distributed learning. ONE \cite{LanZG18} constructs a multi-branch network and assembles the on-the-fly logit information from the branches to enhance the performance on the target network. CLNN \cite{SongC18} proposes multiple generated classifier heads to obtain supplementary information for improving the generalization ability of the target network. KDCL \cite{GuoWWYLHL20} aggregates the outputs of numerous students with different learning capacities to generate high-quality labels for supervision. PCL \cite{WuG21b} integrates online ensemble and collaborative learning into a unified framework. \textit{Unlike these works generating a soft target in the logit space and transferring knowledge between the isomorphic CNN-based models, we introduce a collaborative learning strategy between the heterogeneous CNN-based and ViT-based models. We propose to bidirectionally exchange the reliable knowledge in feature and logit spaces for semantic segmentation}.

\section{The Proposed Approach}
% In this section, we first briefly describe the fundamental elements of online knowledge distillation for segmentation and then introduce the proposed CTCL framework (as illustrated in Figure \ref{fig:famework} (b)). Subsequently, we present details of the proposed HFCL, RBSS， and SBSS, respectively.
\subsection{Overview}
An overview of the proposed framework is depicted in Fig.~\ref{fig:famework}\textcolor{red}{(b)}, which consists of three components: a CNN-based student $f\left(\theta^{\mathrm{C}}\right)$, a ViT-based student $f\left(\theta^{\mathrm{V}}\right)$, and the proposed KD modules. Given an input image set $X$, our objective is to enable $f\left(x; \theta^{\mathrm{V}}\right)$ and $f\left(x; \theta^{\mathrm{C}}\right)$ to learn collaboratively that can assign a pixel-wise label $l \in 1, \ldots, K$ to each pixel $p_{i, j}$ in image $x \in X (x \in \mathbb{R}^{H \times W \times 3})$ more accurately than the student itself, where $H$ and $W$ are the height and width of $x, K$ is the number of categories. To achieve this goal, given specific input $x$, we attain the segmentation prediction maps $\left(P^{\mathrm{C}}\right.$ and $\left.P^{\mathrm{V}}\right)$ and feature representations ( $F^{\mathrm{C}}$ and $\left.F^{\mathrm{V}}\right)$ from the two students $f\left(x; \theta^{\mathrm{C}}\right)$ and $f\left(x; \theta^{\mathrm{V}}\right)$, respectively, which can be formulated as:
{\setlength\abovedisplayskip{1pt}
\setlength\belowdisplayskip{1pt}
\begin{equation*}
 \left(P^{\mathrm{C}}, F^{\mathrm{C}}\right)=f\left(x ;\theta^{\mathrm{C}}\right), \quad\left(P^{\mathrm{V}}, F^{\mathrm{V}}\right)=f\left(x ; \theta^{\mathrm{V}}\right).   
\end{equation*}}

The pixel-wise segmentation loss $\text{CE}(\cdot)$ is based on the cross-entropy (CE) loss with the ground-truth (GT) label :
{\setlength\abovedisplayskip{2pt}
\setlength\belowdisplayskip{2pt}
\begin{equation}
\small
\begin{aligned}
  &\mathcal{L}_{\text{CE}}^{C}=\frac{1}{H \!\times\! W} \sum_{h=1}^{H} \sum_{w=1}^{W} \text{CE}\left(\sigma\left({P^{C}}_{(h, w)}\right), y_{(h, w)}\right), \\
  &\mathcal{L}_{\text{CE}}^{V}=\frac{1}{H \!\times\! W} \sum_{h=1}^{H} \sum_{w=1}^{W} \text{CE}\left(\sigma\left({P^{V}}_{(h, w)}\right), y_{(h, w)}\right).
\end{aligned}
\end{equation}}

Here, $\sigma$ is the softmax function, and $y_{(h, w)}$ denotes the GT of the $(h, w)$-th pixel of image $x$.
Our key ideas are two folds.  
% Firstly, to enable the heterogeneous students to learn collaboratively, we exploit DeepLabv3+ \cite{ChenZPSA18} as a CNN-based student and SegFormer as a ViT-based student for semantic segmentation. 
To compensate for the limitations of CNNs and ViT, we first propose HFD to align the features in the low-layer feature space. Secondly, we propose a BSD module to selectively enable both two students to mimic the region-wise and pixel-wise information from each other. We now describe the technical details. 
% More details are provided in the following sections. 
\subsection{Heterogeneous Feature Distillation (HFD) }
Inspired by the observations that CNNs are hard-coded to attend only locally while ViT does not learn to attend locally in earlier layers~\cite{RaghuUKZD21}, we propose a novel HFD module \textit{to make the students learn the heterogeneous features from each other for complementary knowledge in the low-layer feature space}. Specifically, it is efficiently achieved by aligning the transformed features between CNNs and ViT (see Fig.\ref{fig:famework} \textcolor{red}{(a)}). At the top of Fig.\ref{fig:famework} \textcolor{red}{(a)}, we transfer knowledge from the ViT-based student to the CNN-based student. To match the shapes and channels between the first-layer features $F_{1}^{C}$ of CNN-based student and the first-stage features $F_{1}^{V}$, we utilize a linear transformation ${\Gamma}^{C}_{1}$ which consists of $1\times 1$ convolution (conv) and pooling layers. And $F_{1}^{C}$ is transformed to be $F_{1}^{\hat{C}} = {\Gamma}^{C}_{1}(F_{1}^{C})$.  Then, the second-stage ViT block `Attn' has two inputs: (a) the feature $F_{1}^{V}$ and (b) the transformed feature $F_{1}^{\hat{C}}$ and outputs second-stage feature $F_{2}^{V}=\text{Attn}(F_{1}^{V})$ and $\text{Attn}(F_{1}^{\hat{C}})$. To enable the low-layer features of CNNs to mimic low-layer features of ViT, we align $F_{2}^{V}$ and $\text{Attn} (F_{1}^{\hat{C}})$ by using cosine distance and use the discrepancy to optimize CNNs. Similarly, as shown in the right of figure of Fig. \ref{fig:famework}\textcolor{red}{(a)}, we exploit the linear transformation ${\Gamma}^{V}_{1}$ to match the spatial size of CNNs and ViT. We also utilize a linear transformation ${\Gamma}^{V}_{1}$ comprising of $1\times 1$ convolution (conv) and pooling layers to transform $F_{1}^{V}$ as $F_{1}^{\hat{V}}={\Gamma}^{V}_{1}(F_{1}^{V})$. The second-layer of CNNs MLP takes the transformed features $F_{1}^{\hat{V}}$ and $F_{1}^{C}$ as inputs, and outputs $\text{MLP}(F_{1}^{\hat{V}})$ and $F_{2}^{C}=\text{MLP}(F_{2}^{C})$, correspondingly. Then, aligning these outputs with the cosine distance facilitates ViT-based student to learn from CNN-based student and thus improves the performance of ViT. Finally, we make ViT-based student learn the local feature representations and CNN-based student learn global feature representations by HFD, which is defined as:

% Since the distinct network structures and computing paradigms of two students, the inner high-level features can not be aligned naively. 

% Consequently, we exploit to match the channels and shapes of the low-layer heterogeneous features by using specific layers, including $1\times 1$ convolution (conv) and pooling layers. 

% According to the different characteristics of features from the two students, we utilize features from different layers (stages) in CNNs (ViT). 

% For CNNs, we leverage the first-layer features $F_{1}^{C}$

% Due to the different architectures of CNNs and ViT, we exploit the $1\times 1$ convolution (conv) and pooling layers to match the channels and shapes. 

% In this work, we input the first-layer features $F_{1}^{C}$ of CNNs into the second stage of ViT.

% Then, we attain the outputs of the second stage with transformed the first-layer features as inputs.

% and align the outputs with the outputs of the stage where the first-stage patches are inputs. 

% After this alignment, we optimize CNNs by computing the discrepancy between these two different outputs, which means this approach enables the CNN features to mimic the ViT features. This can facilitate CNNs to learn the global interaction as ViT does. 

% Similarly, we input the first-stage features into CNNs and align the two outputs of CNNs, which can address ViT limitation of local processing in earlier layers. 

{\setlength\abovedisplayskip{2pt}
\setlength\belowdisplayskip{2pt}
\begin{equation}
\small
\begin{aligned}
  &\mathcal{L}_{\text{HFD}}^{C}=cos({\text{Attn}}((F_{1}^{\hat{C}})),F_{2}^{V}), \\
  &\mathcal{L}_{\text{HFD}}^{V}=cos({\text{MLP}}({F_{1}^{\hat{V}}}),F_{2}^{C}),
\end{aligned}
\end{equation}}
here $cos$ is the cosine distance measuring the consistency between CNNs and ViT. $\text{Attn}((F_{1}^{\hat{C}})$ is defined as 
{\setlength\abovedisplayskip{2pt}
\setlength\belowdisplayskip{2pt}
\begin{equation*}
\small
\begin{aligned}
\text{Attn}(F_{1}^{\hat{C}}) =
\operatorname{softmax}\left(\frac{F_{1}^{\hat{C}}W^Q (F_{1}^{\hat{C}}W^K)^{T}}{\sqrt{d}}\right) (F_{1}^{\hat{C}}W^V),
\end{aligned}
\end{equation*}}

where $W^Q$, $K^Q$, and $V^Q$ are the projections of $\text{Attn}$ and $d$ is the number of multi-head in ViT. Similarly, ${\text{MLP}}$ block is the second layer of CNNs.

\subsection{Bidirectional Selective Distillation (BSD)}
\begin{figure}[t]
    \centering
    \includegraphics[width=0.99\columnwidth]{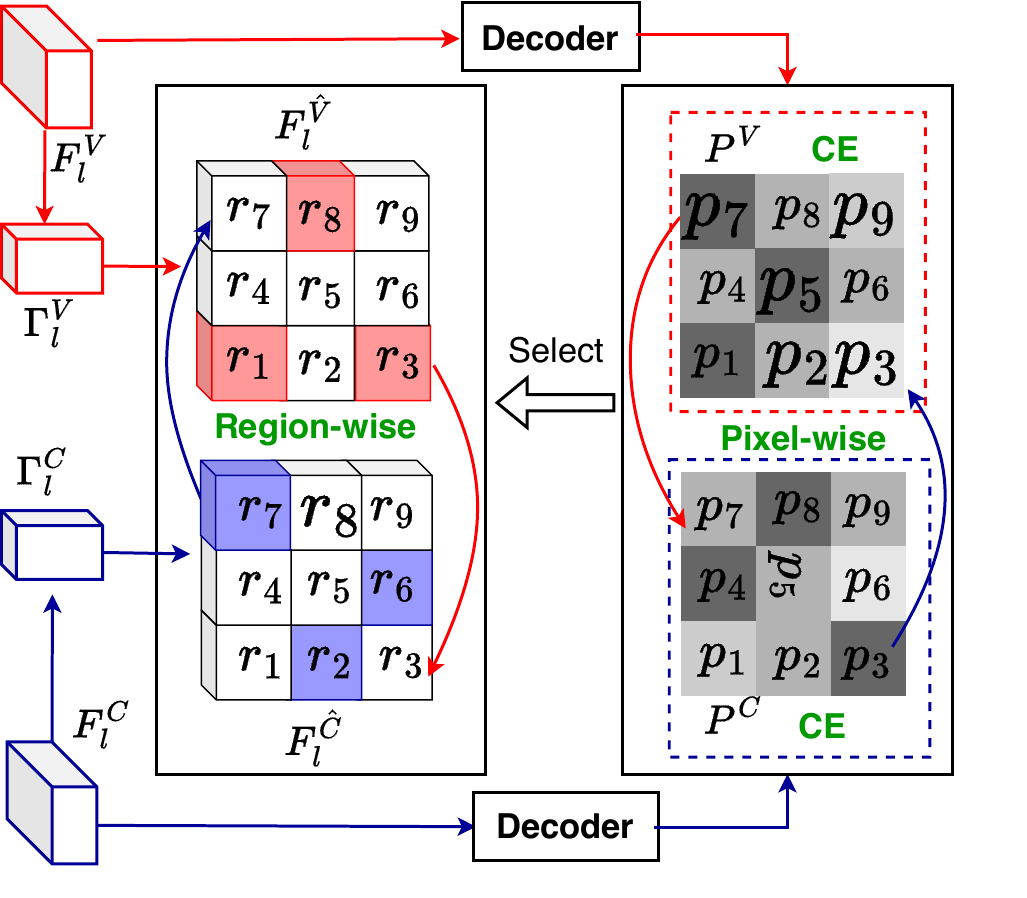}
    \caption{Illustration of the proposed BSD. For region-wise BSD, the colorful cubes mean these regions have more reliable knowledge than the same regions from other students. Then the reliable knowledge is transferred from colorful cubes to white cubes and from more reliable regions to less reliable regions. For pixel-wise BSD, the darker the color of the square, the more accurate the predictions of pixels. Like region-wise, pixel-wise teaches pixels with less accurate predictions from the reliable knowledge of pixels with more accurate predictions. }
    \label{fig:module2}
    \vspace{-10pt}
\end{figure}

Due to the different performance in different regions between the ViT and CNN students, we intend to dynamically select useful knowledge between the two students in the feature space, so as to benefit each other. However, there is a challenging problem: `\textit{how to decide the directions of transferring knowledge for different regions during training}?'. To this end, we propose to manage the directions of KD via combining the predictions and GT labels, where we regard the directions of KD for different regions as a sequential decision-making problem. Consequently, we propose a directional selective distillation (BSD) for \textit{enabling students to learn collaboratively}, as shown in Fig. ~\ref{fig:module2}. Our BSD module transfers knowledge in two aspects. Firstly, the region-wise distillation determines the distillation direction of each region for supervising students in each region. Secondly, the pixel-wise distillation decides which of the prediction knowledge to be transferred in the logit space.  

\subsubsection{Region-wise distillation} 
% Different from previous online knowledge distillation works which only learn collaboratively in the logit space, our work utilizes more rich knowledge in the feature space than the logit space to learn cooperatively. Moreover, 
% This strategy dynamically distills region-wise knowledge between CNN-based and ViT-based students.

Given the last-layer feature $F^{C}_{l} \in \mathbb{R}^{\hat{H} \times \hat{W} \times \hat{D}}$ and the last-stage feature $F^{V}_{l}$, we exploit $1\times1$ conv and pooling layers to transform $F^{C}_{l}$ and $F^{V}_{l}$ for matching the channels and shapes of them. The transformation functions are denoted as $\Gamma_{l}^{C}$ and $\Gamma_{l}^{V}$, respectively. Then $F^{\hat{C}}_{l}=\Gamma_{l}^{C}(F^{C}_{l})$ matches the dimensions of $F^{\hat{V}}_{l}=\Gamma_{l}^{V}(F^{V}_{l})$ and $F^{\hat{C}}_{l} \in R^{\hat{H}\times \hat{W}\times \hat{D}}$. To transfer knowledge from regions between students, we calculate the cross-student region-wise similarity matrix $S_{(\hat{h},\hat{w})} = cos({F^{\hat{C}}_{l}}_{(\hat{h},\hat{w})},(F^{\hat{V}}_{l})_{(\hat{h},\hat{w})}).$ Then, we exploit the cross entropy between the predictions and GT labels to quantify the most reliable knowledge for each region between the students. As shown in Fig.~\ref{fig:module2}, the red grids $\{\textcolor{red}{{r_{1}}}, \textcolor{red}{r_{3}}, \textcolor{red}{r_{8}}\}$ indicate that the knowledge of these regions are more reliable than the white cubes $\{r_{1}, r_{3}, r_{8}\}$, which means the CE losses of these red regions are smaller than blue regions. Then, the direction of KD for these three regions is from red regions to white regions. Specifically, we utilize the matrix $\hat{m}_{(\hat{h},\hat{w})}\in R^{\hat{H} \!\times\! \hat{W}}$ which is 0 or 1 to decides the direction of KD for each region. The value of $\hat{m}_{(\hat{h},\hat{w})}$ is 1 when the CE of this region in CNN-based prediction map is smaller than that of ViT-based prediction map, enabling the knowledge to be transferred from ${F^{\hat{C}}_{l}}_{(\hat{h},\hat{w})}$ to ${F^{\hat{V}}_{l}}_{(\hat{h},\hat{w})}$, and vice versa for $\hat{m}_{(\hat{h},\hat{w})}$=0. Note that for matching the size of ${F^{\hat{C}}_{l}}$ and $P^{C}$, we divide the prediction map into $\hat{H} \times \hat{W}$ size. A $\frac{H}{\hat{H}} \!\times\! \frac{W}{\hat{W}}$ sized prediction map $P^{C}$ at the same location corresponds to one region in ${F^{\hat{C}}_{l}}$. After determining the direction of KD for each region, the two students can exchange reliable region-wise knowledge. Therefore, we weight the similarity matrix $S_{(\hat{h},\hat{w})}$ based on the matrix $\hat{m}_{(\hat{h},\hat{w})}$, which denotes the process of KD from the more reliable regions to these less reliable regions. To achieve this region-wise KD, we propose to minimize the loss function as follows:
{\setlength\abovedisplayskip{2pt}
\setlength\belowdisplayskip{2pt}
\begin{equation}
\small
\begin{aligned}
&\mathcal{L}_{R}^{C} = \frac{1}{\hat{H} \!\times\! \hat{W}\!-\!\hat{M}} \sum_{\hat{h}=1}^{\hat{H}} \sum_{\hat{w}=1}^{\hat{W}} (1-\hat{m}_{(\hat{h},\hat{w})})S_{(\hat{h},\hat{w})},\\ 
&\mathcal{L}_{R}^{V} =  \frac{1}{\hat{M}}\sum_{\hat{h}=1}^{\hat{H}}\sum_{\hat{w}=1}^{\hat{W}}   m_{(\hat{h},\hat{w})}S_{(\hat{h},\hat{w})},
\end{aligned}
\end{equation}}

where $\hat{M} = \sum_{\hat{h}=1}^{\hat{H}} \sum_{\hat{w}=1}^{\hat{W}} \hat{m}_{(\hat{h},\hat{w})}$.

\subsubsection{Pixel-wise distillation}
Previous KD approaches~\cite{WangZJBX20, ShuLGYS21} for semantic segmentation apply the fundamental response-based distillation loss $\mathcal{L}_{\text{KL}}$ for the stable gradient descent optimization: $\mathcal{L}_{\text{KL}}=\frac{1}{H \times W} \sum_{h=1}^{H} \sum_{w=1}^{W} \text{KL} \left({P^{C}}_{(h, w)}\|{P^{V}}_{(h, w)}\right)$, where $\text{KL}(\cdot)$ is the Kullback-Leibler divergence (KL divergence) between two probabilities. However, due to the performance gap, the heterogeneous students have their own strengths in predicting different segmentation categories. Therefore, the pixel-wise distillation \textit{aims to transfer the knowledge of more reliable pixel-wise predictions to less reliable pixel-wise predictions in the logit space}. As shown in Fig.~\ref{fig:module2}, the black squares $\{p_{1},p_{5},p_{7}\}$ of $P^{V}$ means that the CE losses of these pixels are smaller than the gray squares $\{p_{1},p_{5},p_{7}\}$ of $P^{C}$. Therefore, we transfer the reliable knowledge from these more black squares to light black squares. Specifically, we utilize the matrix ${m}_{(h,w)} \in R^{H\!\times\! W}$ which is 0 or 1 to decide the direction of KD for each pixel. The value of ${m}_{(h,w)}$ is 1 when the CE of pixel from CNN-based student is smaller than that of ViT, enabling the knowledge to be transferred from ${p^{C}_{(h,w)}}$ to ${p^{V}_{(h,w)}}$, and vice versa for ${m}_{(h,w)}$=0. Moreover, we use the KL divergence to the effectiveness of transferring knowledge from ViT-based student to CNN-based student: $\text{KL} \left({P^{C}}_{(h, w)}\|{P^{V}}_{(h, w)}\right)$.  After determining the direction of KD for each pixel, the two students can exchange useful pixel-wise knowledge. Therefore, we weight the KL divergence $\text{KL}\left({P^{C}}_{(h, w)}\|{P^{V}}_{(h, w)}\right)$ based on the matrix ${m}_{(h,w)}$, which denotes the process of KD from the more reliable pixels to these less reliable pixels. To achieve pixel-wise distillation, we propose to minimize the loss function as follows:
{\setlength\abovedisplayskip{2pt}
\setlength\belowdisplayskip{2pt}
\begin{equation}
\small
\begin{aligned}
&\mathcal{L}_{P}^{C} = \frac{1}{H \!\times\! W \!-\!M} \sum_{i=1}^{H}\sum_{i=j}^{W} (1\!-\!m_{(h,w)}) \text{KL} \left({P^{C}}_{(h, w)}\|{P^{V}}_{(h, w)}\right), \\
&\mathcal{L}_{P}^{V} =\frac{1}{M} \sum_{i=1}^{H}\sum_{i=j}^{W} m_{(h,w)}\text{KL}\left({P^{V}}_{(h, w)}\|{P^{C}}_{(h, w)}\right),
\end{aligned}
\end{equation}}

where $M = \sum_{h=1}^{H} \sum_{w=1}^{W} m_{(h,w)}$. 
Finally, combining the region-wise and pixel-wise KD losses, the BSD loss is defined as:
{\setlength\abovedisplayskip{2pt}
\setlength\belowdisplayskip{2pt}
\begin{equation}
\small
\begin{aligned}
&\mathcal{L}_{\text{BSD}}^{C} = \mathcal{L}_{R}^{C} + \alpha \mathcal{L}_{P}^{C} , \\
&\mathcal{L}_{\text{BSD}}^{V} =\mathcal{L}_{R}^{V} + \alpha \mathcal{L}_{P}^{V},
\end{aligned}
\end{equation}}
where $\alpha$ is the trade-parameter to balance the region-wise and pixel-wise losses, and $\alpha$ is set to 1.
% $c_{(h,w)}$ is 1 when the cross entropy loss of pixel of CNN-based student is smaller than ViT-based student, enabling the distribution or prediction of ViT-based student to match the CNN-based student, and vice versa for $c_{(h,w)}$. 

% \subsection{Optimization}
% Following the common practical and previous knowledge distillation approaches for semantic segmentation, we also add the fundamental response-based distillation loss $\mathcal{L}^{C}$ and $\mathcal{L}^{V}$ for the stable gradient descent optimization:

% \begin{equation}
% \small
% \begin{aligned}
% &\mathcal{L}^{C}_{KL}=\frac{1}{H \times W} \sum_{h=1}^{H} \sum_{w=1}^{W} KL \left({P^{C}}_{h, w}\|{P^{V}}_{h, w}\right), \\
% &\mathcal{L}^{V}_{KL}=\frac{1}{H \times W} \sum_{h=1}^{H} \sum_{w=1}^{W} KL \left({P^{V}}_{h, w}\|{P^{C}}_{h, w}\right),
% \end{aligned}
% \end{equation}
% where $KL(\cdot)$ is the Kullback-Leibler divergence between two probabilities. 

\subsection{Optimization}

Overall, the objectives of the proposed method for CNN-based and ViT-based students are given as 

\begin{equation}
\small
\begin{aligned}
&\mathcal{L}^{C}=\mathcal{L}^{C}_{\text{CE}}+\beta \mathcal{L}^{C}_{\text{HFD}}+\gamma\mathcal{L}^{C}_{\text{BSD}}, \\
&\mathcal{L}^{V}=\mathcal{L}^{V}_{\text{CE}}+\beta \mathcal{L}^{V}_{\text{HFD}}+\gamma\mathcal{L}^{V}_{\text{BSD}},
\end{aligned}
\end{equation}

where $\beta$ and $\gamma$ are hyperparameters and set to 0.1 and 1, respectively.
% \begin{figure}[t]
%  \centering
% \subcaptionbox{\label{fig:mutual1}}{
% \includegraphics[width=0.46 \linewidth]{Mutual1.pdf}
% }
% \hfill
% \subcaptionbox{\label{fig:mutual2}}{
% \includegraphics[width=0.46 \linewidth]{Mutual2.pdf}
% }
% \caption{Mutual learning between CNNs and ViT.}
%  \label{fig:mutual learning}
% \end{figure}

% \begin{figure}[t]
%     \centering
%     \includegraphics[width=0.95\linewidth]{selective knowledge.pdf}
%     \caption{Selective supervision module}
%     \label{fig:selective supervision module}
% \end{figure}

\section{Experiments and Evaluation}
\subsection{Setup}
\textbf{Datasets.} In this work, we conduct extensive experiments to demonstrate the effectiveness of the proposed method on three public datasets: \textbf{PASCAL VOC 2012} \cite{EveringhamEGWWZ15}, \textbf{Cityscapes} \cite{CordtsORREBFRS16}, and \textbf{CamVid} \cite{BrostowSFC08}. Following previous works \cite{WangZJBX20, ShuLGYS21}, we adopt the augmented \textbf{PASCAL VOC 2012} set \cite{hariharan2011semantic} consisting of 10,582 training and 1,449 validation images with 21 pixel-wise annotated classes. \textbf{Cityscapes} is a dataset for urban scene understanding and consists of 5,000 fine-annotated 1024 $\times$ 2048 images with 19 categories for segmentation. \textbf{CamVid} is another widely used urban scene dataset with 11 classes, such as building, tree, sky, car, road, etc., and the 12th class indicates unlabeled data. It contains 367 training, 101 validation, and 233 testing images of 720 $\times$ 960, where we resize them to 360 $\times$ 480 following previous work.

\begin{table*}[t]
    \centering
    \small 
    \captionsetup{font=small}
    \renewcommand{\tabcolsep}{10.5pt}
    \begin{tabular}{|l|c|c|c|c|c|c|c|}
    \hline
    Dataset&Method& MobileNet&MiT-B1&$\Delta$&ResNet-50&MiT-B2&$\Delta$\\
    \hline\hline
    \multirow{6}{*}{PASCAL VOC 2012} &Vanilla & 67.54 &78.48&0.00& 76.05 &82.03&0.00 \\
    &Offline KD &$67.40_{\textcolor{blue}{-0.14}}$&$78.87_{\textcolor{red}{+0.39}}$&\textcolor{red}{+0.25}&$76.77_{\textcolor{red}{+0.68}}$&$82.19_{\textcolor{red}{+0.16}}$&\textcolor{red}{+0.88}\\
    &DML&$67.43_{\textcolor{blue}{-0.11}}$&$78.76_{\textcolor{red}{+0.28}}$&\textcolor{red}{+0.17}&$76.51_{\textcolor{red}{+0.46}}$&$82.10_{\textcolor{red}{+0.07}}$&\textcolor{red}{+0.53}\\
    &KDCL&$67.41_{\textcolor{blue}{-0.13}}$&$78.76_{\textcolor{red}{+0.28}}$&\textcolor{red}{+0.15}&$76.46_{\textcolor{red}{+0.41}}$&$82.01_{\textcolor{blue}{-0.02}}$&$\textcolor{red}{+0.39}$\\
    &IFVD & $67.70_{\textcolor{red}{+0.16}}$ & $77.61_{\textcolor{blue}{-0.87}}$& \textcolor{blue}{-0.71}&$76.52_{\textcolor{red}{+0.47}}$ & $81.52_{\textcolor{blue}{-0.51}}$ &\textcolor{blue}{-0.03}\\
%    DKD \cite{abs-2203-08679}&69.43&77.46&0.87\\
   & \textbf{Ours}&$\textbf{69.57}_{\textcolor{red}{+2.03}}$&$\textbf{79.17}_{\textcolor{red}{+0.69}}$&\textcolor{red}{\textbf{+2.72}}& $\textbf{76.99}_{\textcolor{red}{+0.94}}$ & $\textbf{82.67}_{\textcolor{red}{+0.64}}$ &$\textcolor{red}{\textbf{+1.58}}$\\
    \hline
    \end{tabular}
    \caption{Comparison with the SoTA KD methods on the \textbf{PASCAL VOC 2012} dataset for our CNN-based (MobileNetV2 and ResNet-50) and ViT-based (MiT-B1 and MiT-B2) students.
% Quantitative results our method with the state-of-the-art KD methods on PASCAL VOC 2012.
}
    \label{tab:Results_VOC}
\end{table*}
\begin{figure*}[t!]
\captionsetup{font=small}
    \centering
    \includegraphics[width=\textwidth,height=230pt]{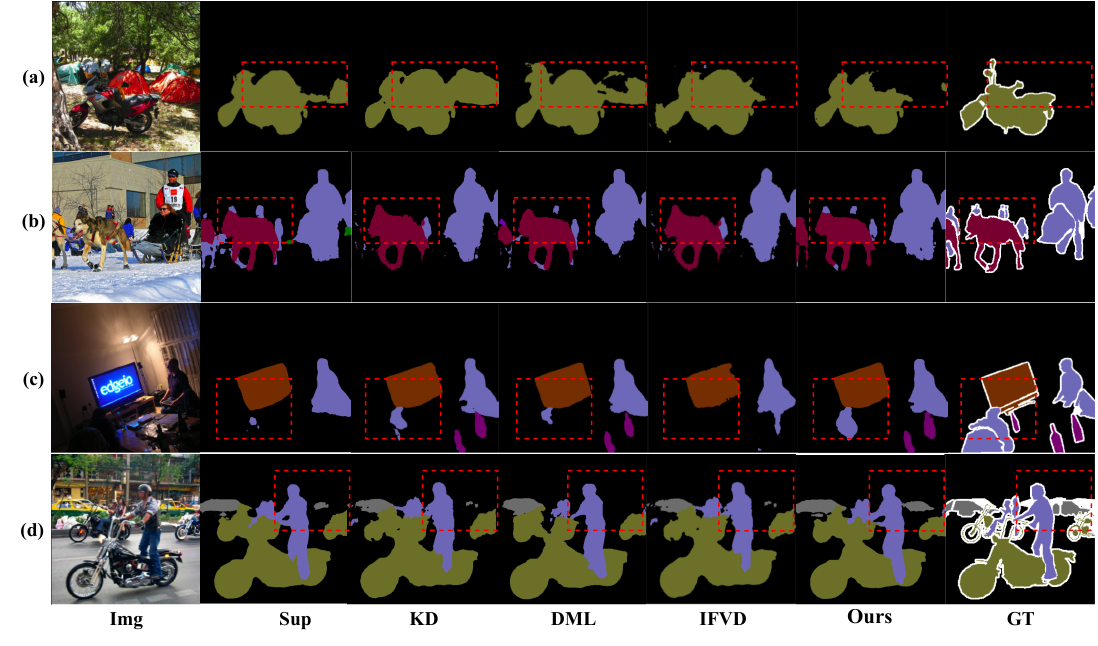}
      \vspace{-12pt}
    \caption{\textbf{Visual results on PASCAL VOC 2012.} (a) MobileNetV2, (b) ResNet-50, (c) MiT-B1 and (d) MiT-B2.}
    \label{cityresults}
\vskip -0.20in
\end{figure*}

\noindent \textbf{Implementation and Evaluation}
We implement our method on PyTorch framework. For CNN-based students, we adopt the widely applied segmentation architecture DeepLabV3+ with encoders of MobileNetV2 and ResNet-50;  for ViT-based students, we utilize the efficient SegFormer with encoders of MiT-B1 and MiT-B2, which have comparable or smaller parameters with their CNN counterparts, respectively. Due to the page limit, we put the details in the supplementary.

In each dataset, CNN-based students are trained by mini-batch stochastic gradient descent (SGD) where the momentum is 0.9, and weight decay is 0.0005; and ViT-based students are trained by AdamW optimizer with a learning rate 0.00006 and weight decay of 0.01. We train Pascal VOC 2012 for 60 epochs with image size 512 $\times$ 512, where the learning rate for CNN-based models is set to 0.0025 and ViT-based to 0.00006. We evaluate the performance by the mean Intersection over Union (mIoU) score and report our results on the validation sets. We use center-crop evaluation for Pascal VOC 2012 and sliding windows evaluation for Cityscapes. We randomly crop images as 512 $\times$ 512 inputs trained with a batch size of 4. It is worth mentioning that for Offline KD and IFVD approaches, the CNN-based and ViT-based students guide each other as teachers. 
\begin{table*}[t]
    \centering
    \small
    \captionsetup{font=small}
    \renewcommand{\tabcolsep}{10.5pt}
    \begin{tabular}{|l|c|c|c|c|c|c|c|}
    \hline
    Dataset&Method& MobileNet&MiT-B1&$\Delta$&ResNet-50&MiT-B2&$\Delta$\\
    \hline\hline
     \multirow{6}{*}{Cityscapes}&Vanilla &73.23&74.95&0.00&76.83&78.77&0.00 \\
    &Offline KD  &$74.11_{\textcolor{red}{+0.84}}$  & $75.50_{\textcolor{red}{+0.55}}$ &\textcolor{red}{+1.43} &$77.68_{\textcolor{red}{+0.85}}$ &$78.84_{\textcolor{red}{+0.07}}$ &\textcolor{red}{+0.92}\\
    &DML &$73.68_{\textcolor{red}{+0.45}}$&$75.13_{\textcolor{red}{+0.18}}$&\textcolor{red}{+0.63}&$77.22_{\textcolor{red}{+0.39}}$&$78.91_{\textcolor{red}{+0.14}}$&\textcolor{red}{+0.53}\\
    &KDCL &$73.41_{\textcolor{red}{+0.28}}$&$75.51_{\textcolor{red}{+0.56}}$&\textcolor{red}{+0.74}&$77.94_{\textcolor{red}{+1.11}}$&$78.81_{\textcolor{red}{+0.04}}$&\textcolor{red}{+1.15}\\
    &IFVD&$73.13_{\textcolor{blue}{-0.10}}$&$75.25_{\textcolor{red}{+0.30}}$ & \textcolor{red}{+0.20} &$77.57_{\textcolor{red}{+0.74}}$ &$78.90_{\textcolor{red}{+0.13}}$&\textcolor{red}{+0.83}\\
    %DKD \cite{abs-2203-08679}&&74.72&&&&\\
    &\textbf{Ours} &  $\textbf{74.42}_{\textcolor{red}{+1.19}}$& $\textbf{75.62}_{\textcolor{red}{+0.37}}$ &\textcolor{red}{\textbf{+1.86}}&$\textbf{78.03}_{\textcolor{red}{+1.20}}$&$\textbf{79.71}_{\textcolor{red}{+0.94}}$&\textcolor{red}{\textbf{+2.14}}\\
    \hline
    \end{tabular}
    \caption{Comparison with the SoTA KD methods on the \textbf{Cityscapes} dataset for our CNN-based (MobileNetV2 and ResNet-50) and ViT-based (MiT-B1 and MiT-B2) students.
% Quantitative results our method with the state-of-the-art KD methods on PASCAL VOC 2012.
}
    \label{tab:Results_Cityscapes}
    \vspace{-5pt}
\end{table*}
\begin{table*}[t]
\captionsetup{font=small}
    \centering
    \small
    \renewcommand{\tabcolsep}{10.5pt}
    \begin{tabular}{|l|c|c|c|c|c|c|c|}
    \hline
    Dataset&Method& MobileNet&MiT-B1&$\Delta$&ResNet-50&MiT-B2&$\Delta$\\
    \hline\hline
      \multirow{6}{*}{CamVid}&  Vanilla &  71.28 & 76.10  &0.00& 73.97 & 77.04 &  0.00 \\
   & Offline KD  &$69.48_{\textcolor{blue}{-1.80}}$  &  $75.76_{\textcolor{blue}{-0.34}}$ &  \textcolor{blue}{-2.14} & $71.90_{\textcolor{blue}{-2.07}}$& $77.29_{\textcolor{red}{+0.21}}$ &\textcolor{blue}{-1.28}  \\
   & DML&$70.73_{\textcolor{blue}{-0.55}}$&$75.85_{\textcolor{blue}{-0.25}}$&\textcolor{blue}{-0.80}&$73.75_{\textcolor{blue}{-0.22}}$&$77.15_{\textcolor{red}{+0.11}}$&\textcolor{blue}{-0.41}\\
    &KDCL &$71.96_{\textcolor{red}{+0.68}}$&$76.40_{\textcolor{red}{+0.30}}$&\textcolor{red}{+0.98}&$73.19_{\textcolor{blue}{-0.78}}$&$77.56_{\textcolor{red}{+0.51}}$&\textcolor{blue}{-0.26}\\
   & IFVD  & $71.08_{\textcolor{blue}{-0.20}}$ & $75.38_{\textcolor{blue}{-0.72}}$ & \textcolor{blue}{-0.92} & $74.22_{\textcolor{red}{+0.25}}$ & $77.25_{\textcolor{red}{+0.21}}$ &\textcolor{red}{+0.46}  \\
   % DKD \cite{abs-2203-08679}&&&&&&\\
    &\textbf{Ours}  & $\textbf{73.09}_{\textcolor{red}{+1.81}}$ & $\textbf{77.04}_{\textcolor{red}{+0.94}}$ & \textcolor{red}{\textbf{+2.75}} &  $\textbf{75.26}_{\textcolor{red}{+1.29}}$   &     $\textbf{78.52}_{\textcolor{red}{+1.48}}$ &  \textcolor{red}{\textbf{+2.77}} \\
    \hline
    \end{tabular}
    \caption{Comparison with the SoTA KD methods on the \textbf{CamVid} dataset for our CNN-based (MobileNetV2 and ResNet-50) and ViT-based (MiT-B1 and MiT-B2) students.
% Quantitative results our method with the state-of-the-art KD methods on PASCAL VOC 2012.
}
\vspace{-5pt}
    \label{tab:Results_Camvid}
\end{table*}

\subsection{Experiments results}
We conduct experiments with two pairs: MobileNetV2 and MiT-B1; ResNet-50 and MiT-B2. It is worth noting that we build two tasks on online KD, which means that students can transfer knowledge from each other. And we compare our proposed method with some SoTA KD methods Offline KD \cite{HintonVD15}, DML \cite{ZhangXHL18}, KDCL \cite{GuoWWYLHL20}, and IFVD \cite{WangZJBX20}. Furthermore, the $\Delta$ is the sum of the performance improvement of each pair compared with Vanilla. 

\noindent\textbf{Results on PASCAL VOC 2012.} We first evaluate the proposed method on the PASCAL VOC 2012 dataset and report the quantitative results in Tab. \ref{tab:Results_VOC}. Our findings show that existing SoTA KD methods designed for isomorphic models have inferior generalization abilities in online KD between CNN and ViT, compared to the vanilla KD method. In contrast, our proposed approach exhibits significantly better performance. Specifically, our method improves the mIoU of MobileNetV2 and MiT-B1 by \textbf{+2.72\%}. In contrast, the prior SoTA KD methods DML and KDCL only achieved a mIoU increment of \textbf{+0.25\%} and \textbf{+0.15\%}, respectively.

Note that the offline KD method IFVD, designed to transfer knowledge between isomorphic models, impedes the performance of online KD between CNN and ViT. This leads to a drop in performance of -0.71\% in mIoU. The first reason mentioned in the introduction causes this outcome. Our proposed method consistently outperforms SoTA KD methods with larger backbone models, ResNet50 and MiT-B2, by achieving a \textbf{+1.58\%} mIoU increment. This result indicates the superiority of our method, which enables students to learn heterogeneous features from each other to acquire complementary knowledge in the feature space.

Fig.~\ref{cityresults} shows the qualitative results and a comparison of the SoTA KD methods on the PASCAL VOC 2012 dataset. Intuitively, the vanilla KD method (3rd column) and DML (4th column) produce unsatisfactory predictions and even erroneous segmentation. In contrast, the results of our method are closer to the ground truth with better segmentation. \textit{\textbf{These outcomes demonstrate the effectiveness and superiority of the proposed BSD module, which selectively distills the reliable region-wise and pixel-wise knowledge.}}

\noindent\textbf{Results on \textbf{Cityscapes}.} 
Tab. \ref{tab:Results_Cityscapes} presents the quantitative results on Cityscapes validation set. Our proposed method consistently outperforms the SoTA KD methods. In comparison with other online KD methods, our method demonstrates a remarkable increase of mIoU by \textbf{+1.86\%}, which is much higher than the online DML's improvement of +0.63\%, while KDCL and IFVD show only +0.74\% and +0.20\% improvements in mIoU, respectively. Our method also outperforms the offline KD methods, Offline KD and IFVD, in terms of segmentation results. Specifically, our method achieves an improvement of \textbf{+2.14\%} in mIoU with the larger backbone ResNet-50 and MiT-B2, while IFVD and KDCL only show \textbf{+0.83\%} and \textbf{+1.15\%} improvements, respectively.
\begin{table}[t!]
\captionsetup{font=small}
  \renewcommand{\tabcolsep}{5pt}
    \centering
    \small
    \begin{tabular}{|l|c|c|c|c|c|}
\hline
$\mathcal{L}_{s}$&$\mathcal{L}_{\text{HFD}}$&$\mathcal{L}_{\text{BSD}}$ & MobileNetV2&MiT-B1&$\Delta$\\
    \hline
    \hline
    \makecell[c]{\cmark}&\makecell[c]{\xmark}&\makecell[c]{\xmark}&67.54 &78.48&0 \\
    \makecell[c]{\cmark}&\makecell[c]{\cmark}&\makecell[c]{\xmark}&$67.89_{\textcolor{red}{+0.35}}$&$78.78_{\textcolor{red}{+0.30}}$&\textcolor{red}{+0.65} \\
    \makecell[c]{\cmark}&\makecell[c]{\xmark}&\makecell[c]{\cmark}&$69.19_{\textcolor{red}{+1.65}}$&$78.91_{\textcolor{red}{+0.43}}$&\textcolor{red}{+2.08} \\
    \makecell[c]{\cmark}&\makecell[c]{\cmark}&\makecell[c]{\cmark} &$69.57_{\textcolor{red}{+2.03}}$&$79.17_{\textcolor{red}{+0.0.69}}$&\textcolor{red}{+2.72}\\
   \hline
    \end{tabular}
    \caption{Ablation of two components of the proposed method evaluated on \textbf{PASCAL VOC 2012}.}
    \vspace{-10pt}
    \label{tab:ablation1}
    \vspace{-5pt}
\end{table}
\begin{table}[t!]
\captionsetup{font=small}
  \renewcommand{\tabcolsep}{10pt}
    \centering
    \small
    \begin{tabular}{|l|c|c|c|c|}
    \hline
    $\mathcal{L}_{R}$&$\mathcal{L}_{P}$& MobileNetV2&MiT-B1&$\Delta$\\
    \hline\hline
    \makecell[c]{\xmark}&\makecell[c]{\xmark}&67.54 &78.48&0 \\
    \makecell[c]{\cmark}&\makecell[c]{\xmark}&$67.92_{\textcolor{red}{+0.38}}$& $78.82_{\textcolor{red}{+0.34}}$&\textcolor{red}{+0.72}\\
    \makecell[c]{\xmark}&\makecell[c]{\cmark}&$68.87_{\textcolor{red}{+1.33}}$&$78.81_{\textcolor{red}{+0.33}}$&\textcolor{red}{+1.66}\\
    \makecell[c]{\cmark}&\makecell[c]{\cmark}&$69.19_{\textcolor{red}{+1.65}}$&$78.91_{\textcolor{red}{+0.43}}$&\textcolor{red}{+2.08}\\
    \hline
    \end{tabular}
    \vspace{-5pt}
    \caption{Ablation of two components of BSD evaluated on \textbf{PASCAL VOC 2012}.}
    \label{tab:ablation2}
    \vspace{-10pt}
\end{table}

\noindent\textbf{Results on CamVid.} Tab. \ref{tab:Results_Camvid} presents a comparison of our proposed method with SoTA KD methods on the CamVid dataset. The results demonstrate the significant performance improvement of ViT-based and CNN-based student models achieved by our method. In contrast to the students without distillation, our method produces a remarkable improvement of \textbf{1.81\%}, \textbf{0.94\%}, \textbf{1.29\%}, and \textbf{1.48\%} in MobileNetV2, MiT-B1, ResNet-50, and MiT-B2, respectively. While most of the previous KD methods show decreased performance on this dataset, our method exhibits better generalization and enables the students to learn collaboratively. Additionally, our method outperforms the compared KD methods, regardless of the choice of different architectures and backbones for the student networks.

\begin{table*}[t]
  \renewcommand{\tabcolsep}{5pt}
  \captionsetup{font=small}
    \centering
    \begin{tabular}{|l|c|c|c|c|c|c|}
    \hline
    Method&MobileNetV2&MiT-B2&$\Delta$&ResNet-50&MiT-B1&$\Delta$\\
    \hline\hline
    Vanilla&67.54&82.03&0.00&76.05&78.48&0.00\\
    \textbf{Ours}&$\textbf{69.21}_{\textcolor{red}{+1.67}}$&$\textbf{82.27}_{\textcolor{red}{+0.24}}$&\textcolor{red}{\textbf{+1.91}}&$\textbf{77.59}_{\textcolor{red}{+1.54}}$&$\textbf{79.56}_{\textcolor{red}{+1.08}}$&\textcolor{red}{\textbf{+2.62}}\\
    \hline
    \end{tabular}
    \vspace{-5pt}
    \caption{Comparison with the Vanilla  methods on the \textbf{PASCAL VOC 2012} dataset for our CNN-based (MobileNetV2 and ResNet-50) and ViT-based (MiT-B1 and MiT-B2) students.}
    \label{tab:coma}
    \vspace{-10pt}
\end{table*}

\begin{table}[t!]
  \renewcommand{\tabcolsep}{6pt}
  \captionsetup{font=small}
  \small
    \centering
    \scalebox{0.78}{
    \begin{tabular}{|c|c|c|c|c|c|}
    \hline
    $\alpha$&0.1&0.5&1.0&2.0&5.0\\
    \hline\hline
     MobileNetV2&$68.14_{\textcolor{red}{+1.41}}$&$68.76_{\textcolor{red}{+1.22}}$&$69.19_{\textcolor{red}{+1.65}}$&$69.47_{\textcolor{red}{+1.93}}$&$70.69_{\textcolor{red}{+3.15}}$\\
    MiT-B1  &$78.94_{\textcolor{red}{+0.46}}$&$78.83_{\textcolor{red}{+0.35}}$&$78.91_{\textcolor{red}{+0.48}}$&$78.55_{\textcolor{red}{+0.07}}$&$78.25_{\textcolor{blue}{-0.23}}$\\
    $\Delta$&\textcolor{red}{1.06} &\textcolor{red}{+1.57}&\textcolor{red}{+2.08}&\textcolor{red}{+2.00}&\textcolor{red}{+2.92}\\
    \hline
    \end{tabular}}
    \caption{Sensitivity of $\alpha$ evaluated on \textbf{PASCAL VOC 2012}.}
    \label{tab:Sensitivity to loss parameter2}
    \vspace{-10pt}
\end{table}

\begin{table}[t!]
  \renewcommand{\tabcolsep}{3.5pt}
    \centering
    \small
    \captionsetup{font=small}
    \begin{tabular}{|c|c|c|c|c|c|c|}
    \hline
    $\beta$&0.05&0.1&0.5&1.0\\
    \hline\hline
    MobileNetV2&$67.60_{\textcolor{red}{+0.06}}$&$67.89_{\textcolor{red}{+0.35}}$&$67.76_{\textcolor{red}{+0.22}}$&$67.44_{\textcolor{blue}{-0.10}}$\\
    MiT-B1 &$78.70_{\textcolor{red}{+0.22}}$&$78.78_{\textcolor{red}{+0.30}}$&$78.69_{\textcolor{red}{+0.21}}$&$78.43_{\textcolor{blue}{-0.05}}$\\
    $\Delta$&\textcolor{red}{+0.28}&\textcolor{red}{+0.65}&\textcolor{red}{+0.43}&\textcolor{blue}{-0.15}\\
    \hline
    \end{tabular}
    \caption{Sensitivity of $\beta$ evaluated on \textbf{PASCAL VOC 2012}.}
    \label{tab:Sensitivity to loss parameter1_beta}
    \vspace{-13pt}
\end{table}
\subsection{Ablation study and analysis}
\noindent\textbf{Effectiveness of the two proposed modules.} We investigate the impact of enabling and disabling the two components of our proposed method on the PASCAL VOC 2012 dataset using MobileNetV2 and MiT-B1. Tab. \ref{tab:ablation1} reports the results of the different student settings. The table shows that both proposed components can enhance the performance of both students, and the selection of reliable knowledge aids better collaborative learning. Specifically, the BSD module improves performance by \textbf{2.08\%}, demonstrating the effectiveness of selecting reliable knowledge to transfer between heterogeneous models.

\noindent\textbf{Effectiveness of the Two Components of BSD.} Tab. \ref{tab:ablation2} demonstrates the effectiveness of the different components in the BSD module. The CNN-based and ViT-based students with the region-wise distillation module achieve results of \textbf{67.92\%} and \textbf{78.42\%}, respectively. The pixel-wise distillation module boosts the improvement to \textbf{2.08\%}, a substantial enhancement to the students' performance. We observe that pixel-wise distillation can enhance CNNs with a small capacity guided by ViT with a larger capacity. However, distilling knowledge from CNNs improves ViT slightly, indicating a further potential for exploration.

\noindent\textbf{Sensitivity of $\alpha$, $\beta$, and $\gamma$.} Tab. \ref{tab:Sensitivity to loss parameter2}, \ref{tab:Sensitivity to loss parameter1_beta}, and \ref{tab:Sensitivity to loss parameter1_gama} report the mIoU(\%) of the students with different ratios of $\alpha$, $\beta$, and $\gamma$ on the PASCAL VOC 2012 dataset. The students' encoders are MobileNetV2 and MiT-B1.

As shown in Tab. \ref{tab:Sensitivity to loss parameter2}, increasing the importance of $\alpha$ significantly improves the performance of CNNs due to the proposed BSD module's transfer of reliable knowledge from ViT to CNN. However, the performance of ViT degrades slightly when $\alpha$=5.0. As our goal is to facilitate the two students' learning from each other, we choose $\alpha$=1.0 as it presents the best trade-off for both students.

From Tab. \ref{tab:Sensitivity to loss parameter1_beta}, it can be seen that aligning the heterogeneous features in the low-layer feature space directly may degrade the students' performance due to the considerable learning capacity gap between CNNs and ViT. Therefore, we set $\beta$ as 0.1 to facilitate the students to learn from each other, which leads to an absolute improvement of \textbf{+0.65\%}, demonstrating that the HFD module enables the students to learn the heterogeneous features from each other for complementary knowledge in the low-layer feature space.

In Tab. \ref{tab:Sensitivity to loss parameter1_gama}, it shows that as $\gamma$ increases, the performance of CNNs is continually improved, but the performance of ViT degrades slightly. Therefore, we set $\gamma$ to 1.0 as it shows the best trade-off for improving the performance of both CNNs and ViT simultaneously. The results prove that our approach is suitable for situations when there is a significant performance gap between homogeneous students.

\noindent\textbf{Students with different performance ability.} To demonstrate the effectiveness of our proposed method, we conduct experiments with two pairs: MobileNetV2 and MiT-B2, and ResNet-50 and MiT-B1. Tab. \ref{tab:coma} shows the quantitative results on the PASCAL VOC 2012 dataset. Compared to the vanilla methods, our proposed method achieves a dramatic increase in mIoU by \textbf{+1.91\% }and \textbf{+2.62\%}, respectively. The results demonstrate the effectiveness of our proposed method in improving the performance of different heterogeneous students with varying performance abilities. \textit{More details of the method can be found in
suppl. material.}

\begin{table}[t!]
  \renewcommand{\tabcolsep}{6pt}
  \captionsetup{font=small}
    \centering
    \small
    \scalebox{0.78}{
    \begin{tabular}{|c|c|c|c|c|c|}
    \hline
    $\gamma$& 0.1&0.5&1.0&2.0&5.0\\
    \hline\hline
     MobileNetV2&$68.57_{\textcolor{red}{+1.03}}$&$68.92_{\textcolor{red}{+1.38}}$&$69.03_{\textcolor{red}{+1.49}}$&$69.08_{\textcolor{red}{+1.54}}$&$70.12_{\textcolor{red}{+2.58}}$\\
    MiT-B1 &$79.10_{\textcolor{red}{+0.62}}$&$78.95_{\textcolor{red}{+0.47}}$&$78.94_{\textcolor{red}{+0.46}}$&$78.60_{\textcolor{red}{+0.12}}$&$78.30_{\textcolor{blue}{-0.18}}$\\
    $\Delta$&\textcolor{red}{+1.65}&\textcolor{red}{+1.85}&\textcolor{red}{+1.95}&\textcolor{red}{+1.66}&\textcolor{red}{+2.40}\\
    \hline
    \end{tabular}}
    \vspace{-5pt}
    \caption{Sensitivity of $\gamma$ evaluated on \textbf{PASCAL VOC 2012}.}
    \label{tab:Sensitivity to loss parameter1_gama}
    \vspace{-10pt}
\end{table}

\section{Conclusion}
This paper presented the \textbf{first} online KD framework to collaboratively learn compact yet effective CNN-based and ViT-based models for semantic segmentation. Specifically, we proposed the heterogeneous feature distillation (HFD) module to improve students' consistency in the low-layer feature space by mimicking heterogeneous features between CNNs and ViT. Then, we also proposed bidirectional selective distillation (BSD) to select reliable region-wise and pixel-wise knowledge to transfer and enable students to learn from each other better. Comparison with the SoTA KD methods for semantic segmentation shows that our proposed method significantly outperforms these SoTA methods by a large margin and demonstrates our proposed method's effectiveness for semantic segmentation. 

\noindent \textbf{Limitation and future work:} 
% Due that students learn reliable knowledge with the supervision from each other,
% %from each other with the guidance of cross entropy, 
% the performance improvement of the high-performance model is slight, and the high-performance model lacks reliable guidance from low-performance models. 
% supervision
Our method has one limitation.
% As we focus on learning compact ViT-based and CNN-based models. 
The cross-model distillation may lead to an unbalanced performance gain in the online KD training process. That is, if one student's knowledge is less instructive, the performance of the other student may be marginally improved. 
% the distillation from low-performance student  degrade the high-performance model's performance.
% Therefore, we will further explore improving the model with a larger by using a smaller-performance heterogeneous model in an online KD framework. 
Therefore, we will explore the online KD between the heterogeneous ViT-based and CNN-based models with more distinct learning capacities.
Moreover, it is promising to extend the proposed collaborative learning paradigm to learn other tasks than semantic segmentation or the cross tasks between depth estimation and semantic segmentation.

\noindent \textbf{Acknowledgement}: This joint paper is supported by Alibaba Cloud, Alibaba Group through Alibaba Innovative Research Program, and the National Natural Science Foundation of China (NSF) under Grant No. NSFC22FYT45. 
{\small
\bibliographystyle{ieee_fullname}
\bibliography{egbib}
}

\end{document}

%%%%%%%%% TITLE
\title{A Good Student is Cooperative and Reliable: CNN-Transformer Collaborative Learning for Semantic Segmentation

-Supplementary-}

\author{Jinjing Zhu$^{1}$
\quad
Yunhao Luo$^{3}$
\quad
Xu Zheng$^{1}$
\quad
Hao Wang $^{4}$
\quad
Lin Wang$^{1,2}$ \thanks{Corresponding author}
\and
\affmark[1] AI Thrust, HKUST(GZ)\quad
\affmark[2] Dept. of CSE, HKUST \quad
\affmark[3] Brown University\quad
\affmark[4] Alibaba Cloud, Alibaba Group\\
\quad
{\tt\footnotesize zhujinjing.hkust@gmail.com, devinluo27@gmail.com, zhengxu128@gmail.com,
cashenry@126.com, linwang@ust.hk}}

\maketitle

\begin{abstract}
Due to the lack of space in the main paper, we provide more details of the proposed method and experimental results in the supplementary material. Sec.\ref{detials} introduces the details of the proposed method. Sec.\ref{parameters} provides the details of the encoders used in this work. Lastly, Sec.\ref{algorithm} provides pseudo algorithm of the proposed method. Sec.\ref{discussion} shows some discussions about our proposed method.
\end{abstract}

\section{Details of the Proposed Method}
\label{detials}

Tab. \ref{tab:size} shows the architecture of MobileNetV2, ResNet-50, MiT-B1, and MiT-B2, respectively. We take the collaborative learning between MobileNetV2 and MiT-B1 as an example and present the details of our proposed method. 
\subsection{Heterogeneous Feature Distillation (HFD)}

The first-layer feature $F_{1}^{V}$ size of MobileNetV2 is 24$\times$128$\times$128 and the first-stage feature $F_{1}^{V}$ size of MiT-B1 is 64$\times$128$\times$128. To match the sizes of features, we utilize the linear transformations ${\Gamma}^{C}_{1}$ and ${\Gamma}^{V}_{1}$ to reshape the sizes of $F_{1}^{C}$ and $F_{1}^{V}$ as 64$\times$128$\times$128 and 24$\times$128$\times$128, respectively. Then, we can use the transformed feature to calculate the HFD loss as follow:

\begin{equation}
\small
\begin{aligned}
  &\mathcal{L}_{\text{HFD}}^{C}=cos({\text{Attn}}((F_{1}^{\hat{C}})),F_{2}^{V}), \\
  &\mathcal{L}_{\text{HFD}}^{V}=cos({\text{MLP}}({F_{1}^{\hat{V}}}),F_{2}^{C}),
\end{aligned}
\end{equation}
where $F_{1}^{\hat{C}}$ and $F_{1}^{\hat{V}}$ is the transformed feature, the shapes of which are 64$\times$128$\times$128 and 24$\times$128$\times$128, respectively. 
\subsection{Region-wise Bidirectional Selective Distillation}
The last-layer feature $F_{l}^{C}$ size of MobileNetV2 is 96$\times$64$\times$64 and the last-stage feature $F_{l}^{V}$ is 512$\times$ 16$\times$16. To match the sizes of features, we exploit the linear transformations ${\Gamma}^{C}_{l}$ and ${\Gamma}^{V}_{l}$ to reshape the sizes of $F_{1}^{C}$ and $F_{1}^{V}$ as 96$\times$16$\times$16 and 96$\times$16$\times$16, respectively. The transformed features are donated as ${F^{\hat{C}}_{l}}$ and ${F^{\hat{V}}_{l}}$, separately.
It is worth noting that the shapes of the predictions are 512$\times$512. To match the sizes of transformed features ${F^{\hat{C}}_{l}}$ (or ${F^{\hat{V}}_{l}}$ )and predictions $P^{C}$ (or $P^{V}$), we divide the prediction map into $16 \times 16$ size. A $\frac{512}{16} \times \frac{512}{16}$ sized prediction map $P^{C}$ (or $P^{V}$) at the same location corresponds to one region in ${F^{\hat{C}}_{l}}$ (or ${F^{\hat{V}}_{l}}$ ). Then we use the sum of cross entropy loss of $\frac{512}{16} \times \frac{512}{16}$ sized prediction map to decide the transferred direction between two students' regions with the same location.
Finally, the region-wise BSD loss is defined as
\begin{equation}
\small
\begin{aligned}
&\mathcal{L}_{R}^{C} = \frac{1}{16 \times 16-\hat{M}} \sum_{\hat{h}=1}^{16} \sum_{\hat{w}=1}^{16} (1-\hat{m}_{(\hat{h},\hat{w})})S_{(\hat{h},\hat{w})},\\ 
&\mathcal{L}_{R}^{V} =  \frac{1}{\hat{M}}\sum_{\hat{h}=1}^{16}\sum_{\hat{w}=1}^{16}   m_{(\hat{h},\hat{w})}S_{(\hat{h},\hat{w})},
\end{aligned}
\end{equation}
where $\hat{M}$ decides the direction of KD for each region  and  calculate the cross-student region-wise similarity matrix $S_{(\hat{h},\hat{w})}$ is the similarity matrix (as introduced in main paper).

\section{Parameters of Encoder}
\label{parameters}
Tab. \ref{tab:parameter} shows the parameters of encoder for different methods. For CNN-based students, we adopt the famous segmentation architecture DeepLabV3+ with encoders of MobileNetV2 and ResNet-50; for ViT-based students, we utilize the lightweight SegFormer with encoders of MiT-B1 and MiT-B2, which have comparable or smaller parameters with their CNN counterparts, respectively.
\section{Algorithm}
\label{algorithm}
The pseudo algorithm of the proposed method is shown in Algorithm. \ref{alg}.

\begin{table*}[t]
  \renewcommand{\tabcolsep}{5pt}
    \centering
    \begin{tabular}{l|c|c|c|c}
    \toprule
    Layer of MobileNetV2&First-layer $F_{1}^{C}$&Second-layer $F_{2}^{C}$&Third-layer&Last-layer $F_{l}^{C}$\\
    Output Size&24$\times$128$\times$128&32$\times$64$\times$64&64$\times$64$\times$64&96$\times$64$\times$64\\
    \midrule
    Layer of ResNet-50&First-layer $F_{1}^{C}$&Second-layer $F_{2}^{C}$&Third-layer&Last-layer $F_{l}^{C}$\\
    Output Size&256$\times$128$\times$128&512$\times$64$\times$64&1024$\times$32$\times$32&2048$\times$32$\times$32\\
    \midrule
    Stage of MiT-B1&First-stage $F_{1}^{V}$&Second-stage $F_{2}^{V}$&Third-stage&Last-stage $F_{l}^{V}$\\
    Output Size&64$\times$128$\times$128&128$\times$64$\times$64&320$\times$32$\times$32&512$\times$ 16$\times$16\\
    \midrule
    Stage of MiT-B2&First-stage $F_{1}^{V}$&Second-stage $F_{2}^{V}$&Third-stage&Last-stage $F_{l}^{V}$\\
    Output Size&64$\times$128$\times$128&128$\times$64$\times$64&320$\times$32$\times$32&512$\times$ 16$\times$16\\
    \bottomrule
    
    \end{tabular}
    \caption{Output size of each layer (stage) of different encoders. }
    \label{tab:size}
\end{table*}
\begin{table}[t]
    \centering
    \begin{tabular}{c|c|c}
    \toprule
    $Method$&Encoder&Parameters(M)\\
    \midrule
    DeepLabV3+&MobileNetV2&15.4\\
    SegFormer&MiT-B1&13.7\\
    DeepLabV3+&ResNet-50&43.7\\
    SegFormer&MiT-B2&27.5\\
    \bottomrule
    \end{tabular}
    \caption{The Parameters of methods with different encoder.}
    \label{tab:parameter}
\end{table}

\begin{algorithm}[]
	\caption{The Proposed framework} 
	\label{alg} 
	\begin{algorithmic}[1]
	    \STATE \textbf{Input}: $\{X,Y\}$; max iterations: $T$
	    \\ model: $f(X,\theta^{C})$, $f(X,\theta^V)$;
	    \STATE  \textbf{Initialization}: Set $\theta^C$ and $\theta^V$;
	    \FOR{for t $\xleftarrow[]{}$ 1 to $T$}
    	    \STATE Attain the segmentation prediction maps and feature representations for each student, respectively:
    	    \small $ \left(P^{\mathrm{C}}, F^{\mathrm{C}}\right)=f\left(X ;\theta^{\mathrm{C}}\right), \quad\left(P^{\mathrm{V}}, F^{\mathrm{V}}\right)=f\left(X ; \theta^{\mathrm{V}}\right);$
    	    \STATE Compute the pixel-wise segmentation loss for each student:
    	    
    	    $\mathcal{L}_{CE}^{C}=\frac{1}{H \times W} \sum_{h=1}^{H} \sum_{w=1}^{W} CE\left(\sigma\left({P^{C}}_{(h, w)}\right), y_{(h, w)}\right)$, \\
            $\mathcal{L}_{CE}^{V}=\frac{1}{H \times W} \sum_{h=1}^{H} \sum_{w=1}^{W} C E\left(\sigma\left({P^{V}}_{(h, w)}\right), y_{(h, w)}\right);$
            \STATE Compute the HFD loss for each student:
            
            $\mathcal{L}_{\text{HFD}}^{C}=cos({Attn}((F_{1}^{\hat{C}})),F_{2}^{V})$, \\
            $\mathcal{L}_{\text{HFD}}^{V}=cos({MLP}({F_{1}^{\hat{V}}}),F_{2}^{C});$
            \STATE Compute the region-wise BSD loss for each student:
            
            $\mathcal{L}_{R}^{C} = \frac{1}{\hat{H} \times \hat{W}-\hat{M}} \sum_{\hat{h}=1}^{\hat{H}} \sum_{\hat{w}=1}^{\hat{W}} (1-\hat{m}_{(\hat{h},\hat{w})})S_{(\hat{h},\hat{w})}$,\\ 
           $\mathcal{L}_{R}^{V} =  \frac{1}{\hat{M}}\sum_{\hat{h}=1}^{\hat{H}}\sum_{\hat{w}=1}^{\hat{W}}   m_{(\hat{h},\hat{w})}S_{(\hat{h},\hat{w})}$;
           \STATE Compute the pixel-wise BSD loss for each student:
           $\mathcal{L}_{\text{BSD}}^{C} = \mathcal{L}_{R}^{C} + \alpha \mathcal{L}_{P}^{C}$ , \\
          $\mathcal{L}_{\text{BSD}}^{V} =\mathcal{L}_{R}^{V} + \alpha \mathcal{L}_{P}^{V};$
          \STATE Compute the total objective for each student:
          
          $\mathcal{L}^{C}=\mathcal{L}^{C}_{CE}+\beta \mathcal{L}^{C}_{\text{HFD}}+\gamma\mathcal{L}^{C}_{\text{BSD}},$ \\
         $\mathcal{L}^{V}=\mathcal{L}^{C}_{CE}+\beta \mathcal{L}^{V}_{\text{HFD}}+\gamma\mathcal{L}^{V}_{\text{BSD}}.$
         \STATE Back propagation for $\mathcal{L}^{C}$ and $\mathcal{L}^{V}$;
         \STATE Update the students $\theta^C$ and $\theta^{V}$ with $\mathcal{L}^{C}$ and $\mathcal{L}^{V}$, respectively.
    	 \ENDFOR
    % 	 \eindent
	    \STATE  \textbf{return}  $\theta^C$ and $\theta^{V}$.
	    \STATE  \textbf{End}.
	\end{algorithmic} 
\end{algorithm}

\section{Discussion}
\label{discussion}

\subsection{Intuition of BSD}
The design of BSD is one of the critical contributions of this paper as it facilitates the two students to collaboratively learn reliable knowledge from each other and the knowledge is transferred bidirectionally. Due to the different performance at different regions between the ViT and CNN students, we intend to dynamically select reliable knowledge between the two students in the feature space, so as to benefit each other. However, there is a challenging problem: ‘how to decide the directions of transferring knowledge fro different regions during training?’ To this end, we propose to manage the directions of KD via combining the predictions and GT labels, where we regard the directions of KD for different regions as a sequential decision making problem. Consequently, we propose a directional selective distillation (BSD) for enabling students to learn collaboratively. As the principle of collaborative learning requires bidirectional knowledge transfer, BSD should be “bidirectional” to enable CNNs to learn from ViT while ViT learns from CNNs. Our key idea is “selective” due to the considerable model size gap and learning capacity gap between CNNs and ViT. The reasons causing the gaps are 1): The discrepancies in features and predictions between CNNs and ViT caused by the distinct computing paradigms make it challenging to do online KD. 2): These methods only transfer the knowledge in logit space; however, there is more reliable and efficient knowledge in the features extracted by both models. 3) There are considerable model size gap and learning capacity gap between CNNs and ViT.
\begin{table*}[t]
  \renewcommand{\tabcolsep}{5pt}
  \captionsetup{font=small}
    \centering
    \begin{tabular}{|l|c|c|c|c|c|c|}
    \hline
    Method&MobileNetV2&MiT-B2&$\Delta$&ResNet-50&MiT-B1&$\Delta$\\
    \hline\hline
    Vanilla&67.54&82.03&0.00&76.05&78.48&0.00\\
    \textbf{Ours}&$\textbf{69.21}_{\textcolor{red}{+1.67}}$&$\textbf{82.27}_{\textcolor{red}{+0.24}}$&\textcolor{red}{\textbf{+1.91}}&$\textbf{77.59}_{\textcolor{red}{+1.54}}$&$\textbf{79.56}_{\textcolor{red}{+1.08}}$&\textcolor{red}{\textbf{+2.62}}\\
    \hline
    \end{tabular}
    \vspace{-5pt}
    \caption{Comparison with the Vanilla  methods on the \textbf{PASCAL VOC 2012} dataset for our CNN-based (MobileNetV2 and ResNet-50) and ViT-based (MiT-B1 and MiT-B2) students.}
    \label{tab:coma}
    \vspace{-10pt}
\end{table*}
\subsection{Intuition of HFD}

We make students learn the heterogeneous features from each other in the first-layer feature space and align these features in the second layer. That is, we input the transformed features into the second layer and then align the outputs instead of directly aligning features of the first layer. This way, it can make both students learn the global and local features in the first-layer space.

\subsection{Selection of Layers}

We use the first-layer features as low-layer features of CNNs and ViT are less distinct and heterogeneous, making CNNs and ViT learn from each other more effectively. Moreover, due to the different computing paradigms and learning capacities of CNNs and ViT, aligning high-layer features is less approachable and practical. Lastly, aligning multiple low-layer features lead to an increase in the computation cost. Tab. \ref{tab:coma} in the paper shows the effectiveness of our proposed method between heterogeneous students with different performance abilities.

\subsection{About MLP or Attn in HFD Module}
MLP consisting of convolutional layers extracts the local semantic features, and Attn consisting of a self-attention module extracts the global semantic features. Therefore, after inputting the local features into Attn or inputting the global features into MLP, these output features are comparable. As such, we use cosine similarity to measure the similarity of these features and enable students to learn from each other in the low-layer space.

\subsection{About Operations in Eq.\textcolor{red}{2}}
Attn updates the first-layer features of CNNs, while MLP updates the first-layer features of ViT. However, if we apply Attn operation in `CNNs to ViT” and MLP in “ViT to CNNs', Attn operation can optimize the first two layers of ViT while MLP operation can optimize the first two layers of CNNs. Both approaches can facilitate collaborative learning between CNNs and ViT but optimizing the first two layers increases computation cost.

\subsection{About ViT-ViT setting}
ViT is not absolutely better while CNN still matters; therefore, we explore to take full advantage of CNN and ViT while compensating for their limitations. Moreover, in Tab.\ref{tab:vit-vit}, our method demonstrates superior performance compared to previous studies in ViT-ViT setting.

\subsection{Results on ADE-20K:}
The effectiveness of our method is further demonstrated by the results obtained on the more challenging ADE-20K dataset, as shown in Tab.~\ref{tab:vit-vit}. The results will be included in the final version. 

\subsection{Distillation on hybrid network:} 
We explore the potential of our framework between the CNN-based (ViT-based) and hybrid network-based students, to further demonstrate its effectiveness in Tab.~\ref{tab:Results_camvid_resnet50}. The significant improvements \textbf{+7.59\%} and \textbf{+5.45\%} underscore the effectiveness and practicality of employing our proposed methodology within hybrid network architectures. 

\begin{table*}[t]
    \centering
    % \footnotesize
    \begin{tabular}{|l|c|c|c|c|c|c|c|c|c|}
    \hline
    Method&& MiT-B1&MiT-B2&$\Delta$&& MobileNet&MiT-B1&$\Delta$\\
    \hline\hline
    Vanilla &\multirow{5}{*}{\rotatebox{90}{\textbf{CamVid}}}& 76.26&77.76&0.00&\multirow{5}{*}{\rotatebox{90}{\textbf{ADE-20K}}}&22.53&40.07&0.00 \\
    DML &&75.84&77.40&-0.78&&22.02&40.12&-0.46\\
    KDCL&&76.61&77.55&+0.14&&22.16&41.62&+1.18\\
    IFVD&&76.43&77.45&-0.14&&21.42&40.64&-0.54\\
    Ours&&\textbf{77.89}&\textbf{78.01}&\textcolor{red}{\textbf{+1.88}}&&\textbf{26.47}&\textbf{42.28}&\textcolor{red}{\textbf{+6.15}}\\
    \hline
    \end{tabular}%
    \caption{Comparison on the \textbf{CamVid} for MiT-B2 and MiT-B2 students, and  \textbf{ADE-20K} for MobileNetV2 and MiT-B1 students.
}
    \label{tab:vit-vit}
\end{table*}

\begin{table*}[t]
    \centering
    \begin{tabular}{|l|c|c|c||c|c|c|c|}
    \hline
    Method& ResNet-50&MaxViT&$\Delta$& MiT-B2&MaxViT&$\Delta$\\
    \hline\hline
    Vanilla &58.12&61.89&0.00&77.76&61.89&0.00\\
    DML &59.07&63.80&+2.86&77.09&60.61&-1.95\\
    KDCL&58.64&61.61&+0.24&77.49&63.26&+1.10\\
    IFVD&59.69&62.01&+1.69&77.08&63.10&+0.53\\
    Ours&\textbf{62.13}&\textbf{65.47}&\textcolor{red}{\textbf{+7.59}}&\textbf{77.96}&\textbf{67.14}&\textcolor{red}{\textbf{+5.45}}\\
    \hline
    \end{tabular}
    \caption{Comparison on the \textbf{CamVid} for ResNet-50(MiT-B2) and MaxViT students.
}
    \label{tab:Results_camvid_resnet50}
\end{table*}
\begin{figure*}[t]
    \centering
    \includegraphics[width=0.95\textwidth]{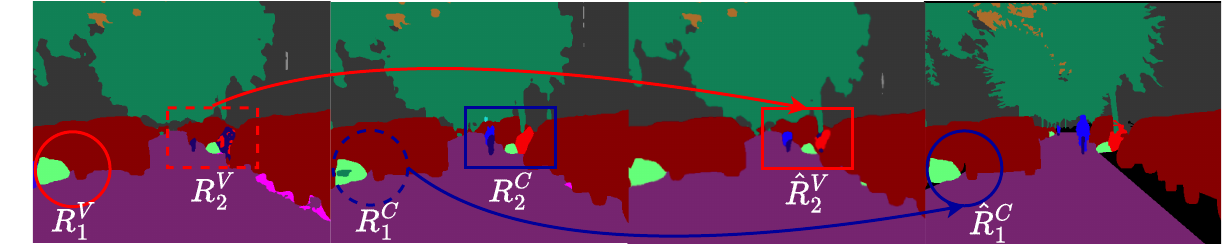}
    \caption{CNN and ViT learns collaboratively by exchanging reliable knowledge.}
    \label{fig:bsd_rel}
\end{figure*}
\subsection{About the motivation}
We argue that CNN is undoubtedly necessary for our problem setting. \textcolor{red}{\textbf{\textcircled{1}}} As ViT is notoriously impeded by limitations, such as the lack of certain inductive biases and poor performance on small-scale datasets; while CNN excels at capturing local features although CNN may underperform ViT on large-scale datasets. Therefore,  ViT is not absolutely better while CNN still matters, and it is promising to take full advantage of CNN and ViT while compensating for their limitations. From this new perspective, prior arts [1,2] adopting the CNN for an auxiliary purpose, are less optimal and intuitive. So, our motivation is reasonable and novel. Our key idea is to simultaneously learn compact yet effective CNN-based and ViT-based models by selecting and exchanging reliable knowledge between them for semantic segmentation. 
 \textcolor{red}{\textbf{\textcircled{2}}} Although `ViT is shown to have higher upper bounds than CNN', we observe in Figs.~\textcolor{red}{1}(b) and ~\textcolor{red}{4} that ViTs may exhibit less accurate segmentation results in certain regions compared to CNNs within the same image. To address this, we introduce BSD to compensate for students' weaknesses in region-wise and pixel-wise levels. We further demonstrate the effectiveness of our proposed method in collaborative learning between CNN-based (or ViT-based) and hybrid network-based students by conducting experiments as shown in Tab.~\ref{tab:Results_camvid_resnet50}.

\subsection{About 'reliable` knowledge in BSD}

Here, `reliable` does not indicate `regions', but \textit{indicates better predictions with relatively higher segmentation accuracy} (See Fig.~\ref{fig:bsd_rel}). Predictions in region $R_{1}^V$ ($R_{2}^C$) of ViT (CNN) is more reliable compared with predictions in region $R_{1}^C$ ($R_{2}^V$) of CNN (ViT). Then we utilize BSD to enable $R_{1}^C$ ($R_{2}^V$) to learn from $R_{1}^V$ ($R_{2}^C$). Finally, we obtain more accurate region predictions $\hat{R}_{1}^C$ ($\hat{R}_{2}^V$). \textit{BSD enables students to learn collaboratively and guarantees the correctness and consistency of soft label}. Qualitative results are in Tabs. \textcolor{red}{4, 5, 7}, and \textcolor{red}{9} (in main paper), and visualized results in Fig.~\textcolor{red}{4} specifically highlight the effectiveness of BSD.

{\small
\bibliographystyle{ieee_fullname}
\bibliography{egbib}
}